\documentclass[runningheads]{llncs}
\usepackage[T1]{fontenc}
\usepackage{multirow}
\usepackage{graphicx}
\usepackage{bbding}
\usepackage[dvipsnames]{xcolor}
\definecolor{myyellow}{HTML}{FFC000}
\definecolor{mygreen}{HTML}{00B050}
\definecolor{myblue}{HTML}{0070C0}
\usepackage{url}

\usepackage{graphicx}
\begin{document}
\title{You Don’t Have to Be Perfect to Be Amazing: Unveil the Utility of Synthetic Images}
\titlerunning{Synthetic Image Measurements}

\author{Xiaodan Xing\inst{1} \and
Federico Felder\inst{1, 2}\and
Yang Nan\inst{1}\and
Simon Walsh \inst{1, 2}  \and
Guang Yang\inst{1,2, 3} \Envelope}
% index{Xing, Xiaodan}
% index{Felder, Federico}
% index{Yang, Nan} 
% index{Walsh, Simon} 
% index{Yang, Guang} 

\authorrunning{Xing et al.}

\institute{National Heart and Lung Institute, Imperial College London, London, UK \\
\email{g.yang@imperial.ac.uk}
\and
Royal Brompton Hospital, London, UK
\and
Department of Biomedical and Engineering, Imperial College London, London, UK\\
\vspace{-1em}}

\maketitle

\begin{abstract}
Synthetic images generated from deep generative models have the potential to address data scarcity and data privacy issues. The selection of synthesis models is mostly based on image quality measurements, and most researchers favor synthetic images that produce realistic images, i.e., images with good fidelity scores, such as low Fréchet Inception Distance (FID) and high Peak Signal-To-Noise Ratio (PSNR). However, the quality of synthetic images is not limited to fidelity, and a wide spectrum of metrics should be evaluated to comprehensively measure the quality of synthetic images. In addition, quality metrics are not truthful predictors of the utility of synthetic images, and the relations between these evaluation metrics are not yet clear. In this work, we have established a comprehensive set of evaluators for synthetic images, including fidelity, variety, privacy, and utility. By analyzing more than 100k chest X-ray images and their synthetic copies, we have demonstrated that there is an inevitable trade-off between synthetic image fidelity, variety, and privacy. In addition, we have empirically demonstrated that the utility score does not require images with both high fidelity and high variety. For intra- and cross-task data augmentation, mode-collapsed images and low-fidelity images can still demonstrate high utility. Finally, our experiments have also showed that it is possible to produce images with both high utility and privacy, which can provide a strong rationale for the use of deep generative models in privacy-preserving applications. Our study can shore up comprehensive guidance for the evaluation of synthetic images and elicit further developments for utility-aware deep generative models in medical image synthesis.

\keywords{Synthetic Data Augmentation \and Medical Image Synthesis.}
\end{abstract}
\section{Introduction}
\begin{figure}
    \centering
    \includegraphics[width=\textwidth]{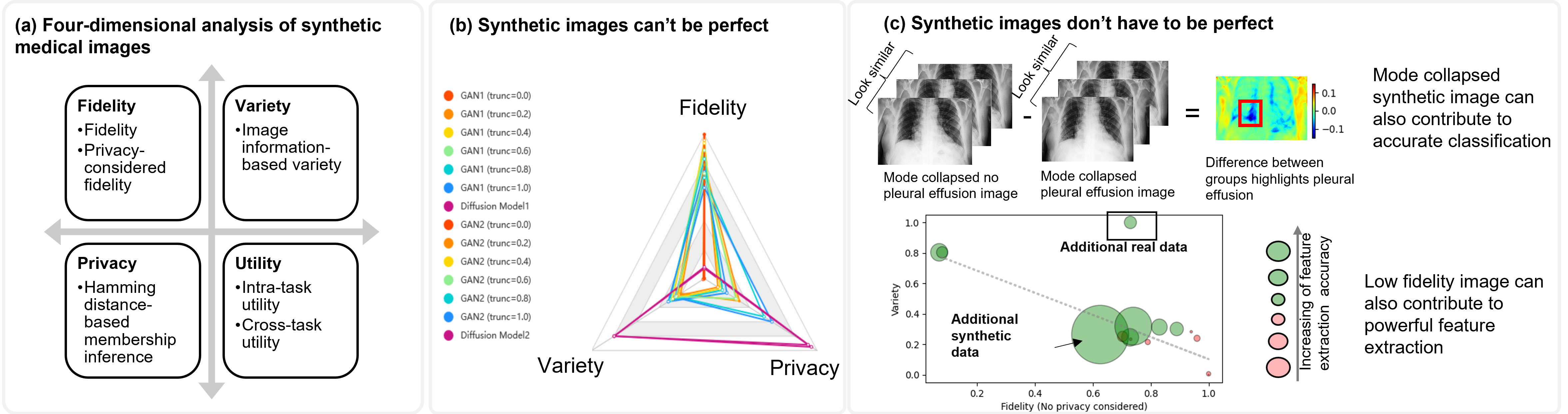}
     \vspace{-7mm}
    \caption{The major contribution (a) and conclusions (b,c) of this work. Our study proposes a comprehensive and disentangled four-dimensional framework for synthetic medical images, incorporating fidelity, variety, privacy, and utility. We conduct extensive experiments to investigate the interplay between these dimensions and identify a set of best practices (c) for selecting synthetic models for downstream tasks. }
    \label{fig:teaser}
    \vspace{-6mm}
\end{figure}

In 2002, SMOTE \cite{chawla2002smote} was proposed to generate synthetic samples to increase the accuracy of classification tasks. Since then, synthetic data has emerged as a promising solution for addressing the data scarcity problem by generating additional training data to supplement the limited real-world datasets. In addition, the potential of synthetic data for privacy preservation has led to the development of generative deep learning models that have shown promise in producing high-quality synthetic data which maintain the statistical properties of the original data while preserving the privacy of individuals in the training data.\\
\indent Deep learning practitioners have been using various metrics to evaluate synthetic images, including Fréchet Inception Distance (FID) \cite{heusel2017gans}, Inception Score (IS) \cite{salimans2016improved}, precision, and recall \cite{lucic2018gans}. However, these measurements are entangled, i.e., they are only able to measure image quality holistically, not a specific aspect. For example, FID is defined by  
\begin{equation}
    {\rm FID}=|(|\mu_1-\mu_2 |)|^2+{\rm Tr}(\Sigma_1+\Sigma_2-2\sqrt{(\Sigma_1 \Sigma_2}).	
\end{equation}
Here, where $\mu_1$, $\mu_2$ $\Sigma_1$, and $\Sigma_2$ are the mean vectors and covariance matrices of the feature representations of two sets of images. A high difference between the image diversity ($\Sigma_1$ and $\Sigma_2$) also leads to a high FID score, which further complicates fidelity evaluation. Another entangled fidelity evaluation is the precision. As is shown in Fig. \ref{fig:function} (a), a high-precision matrix cannot identify non-authentic synthetic images which are copies of real data. Thus, a high-precision matrix can either be caused by high fidelity, or a high privacy breach. \\
\indent When evaluating these entangled metrics, it is difficult to find the true weakness and strengths of synthetic models. In addition, large-scale experiments are currently the only way to measure the utility of synthetic data. The confusion of evaluation metrics and this time and resource-consuming evaluation of synthetic data utility increase the expenses of synthetic model selection and hinder the real-world application of synthetic data. \\
\indent In this study, we aim to provide a set of evaluation metrics that are mathematically disentangled and measure the potential correlation between different aspects of the synthetic image as in Fig. \ref{fig:teaser} (a). Then, we aim to analyze the predictive ability of these proposed metrics to image utilities for different downstream tasks and provide a set of best practices for selecting synthetic models in various clinical scenarios. We compare two state-of-the-art deep generative models with different parameters using a large open-access X-ray dataset that contains more than 100k data  \cite{irvin2019chexpert}. Through our experiments, we empirically show the negative correlations among synthetic image fidelity, variety, and privacy (Fig. \ref{fig:teaser} (b)). After analyzing their impacts on downstream tasks, we discovered that the common problems in data synthesis, i.e., mode collapse and low fidelity, can sometimes be a merit according to the various motivations of different downstream tasks. \\
\indent Overall, our study contributes new insights into the use of synthetic data for medical image analysis and provides a more objective and reproducible approach to evaluating synthetic data quality. By addressing these fundamental questions, our work provides a valuable foundation for future research and practical applications of synthetic data in medical imaging.

\section{Deep Generative Models and Evaluation Metrics}
In this study, we conducted an empirical evaluation using two state-of-the-art deep generative models: StyleGAN2, which has brought new standards for generative modeling regarding image quality \cite{sauer2022stylegan} and Latent Diffusion Models (LDM) \cite{rombach2022high}. We proposed an analysis framework for synthetic images based on four key dimensions: fidelity, variety, utility, and privacy. Manual evaluation of synthetic image fidelity typically involves human experts assessing whether synthetic images appear realistic. However, this evaluation can be subjective and have high intra-observer variance. While many algorithms can be used to measure synthetic image quality, most of them are designed to capture more than one of the four key dimensions. \\
\begin{figure}
    \centering
    \vspace{-8mm}
    \includegraphics[width=\textwidth]{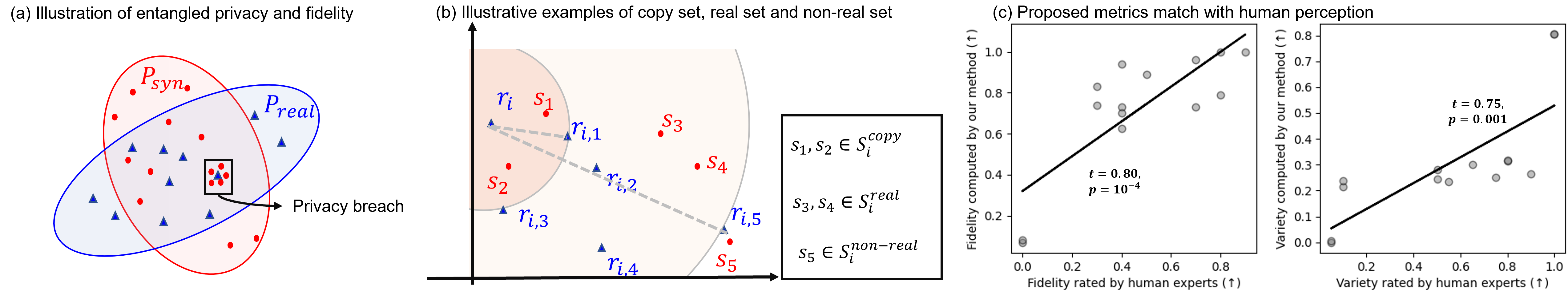}
    \vspace{-6mm}
    \caption{Illustrative examples of entangled privacy and fidelity (a), our proposed algorithms to adjust the fidelity computation (b) and their validity measured by Pearson's correlation analysis (c). In (b), we selected the 5th-nearest neighbor to compute the fidelity, i.e., $k=5$. The human experts were two radiologists with one and two years of experience, respectively.}
    \label{fig:function}
    \vspace{-6mm}
\end{figure}

\indent With this motivation, we aimed to redefine the conventional evaluation metrics of synthetic images and disentangle them into four independent properties. In this study, we employed a metric-based membership inference attack in an unsupervised manner to evaluate the privacy of our model. We presume that synthetic records should have some similarity with the records used to generate them \cite{chen2020ganleaks}. Since differential privacy settings could significantly reduce image fidelity, we chose not to perform differential privacy analysis in this study.\\

 \noindent \textbf{Fidelity and privacy.} Considering we have $M$ real image points and $N$ synthetic image points. For each real image $\{r_i|i\in[0,M)\}$, we select its $k$ nearest neighbors $\{r_{i,1},r_{i,2},…,r_{i,k}\}$ according to the similarity $D(r_i,r_j)$ of these images in the image space. Then, as shown in Fig. \ref{fig:function} (b), we split the synthetic dataset into three sets according to these $k$ nearest neighbors:
\begin{itemize}

\item Copy set $S_i^{\rm copy}$: Synthetic sets that are realistic but are also copies of this real image $r_i$. Synthetic images will belong to this set if this synthetic image is closer to $r_i$ than any other real images, i.e., $D(s_j,r_i) < D(r_i,r_{i,1})$. 
\item Real set $S_i^{\rm real}$: Synthetic sets that are realistic and are not copies of this real image $r_i$. Synthetic images will belong to this set if this synthetic image is in the $k$th-nearest neighbor of $r_i$, i.e., $D(s_j,r_i)\in[(D(r_i,r_{i,1} ),D(r_i,r_{i,k})]$
\item Non-real set $S_i^{\rm non-real}$: Synthetic sets that are not realistic compared to $r_i$. Synthetic images will belong to this set if this synthetic image is not the $k$th-nearest neighbor of $r_i$, i.e., $D(s_j,r_i )>D(r_i,r_{i,k})$
\end{itemize}
We can compute the privacy preservation score of a single synthetic image ${s_j|j\in[0,N)]}$ with the definition of three sets. If the synthetic data $p_j$ is in the copy set of any real data, the privacy protection ability of this $p_j$ is 0, i.e.,
\begin{equation}
p_j=
\left\{ 
  \begin{array}{ c l }
    0 & \quad \textrm{if } \exists i\in [0,M) \to s_j\in S_i^{\rm copy} \\
    1                 & \quad \textrm{otherwise}
  \end{array}
\right. .
\end{equation}
For the synthetic data, the overall privacy protection ability $P\in[0,1]$ was then defined by 
\begin{equation}
    P=\frac{\sum_0^{N}p_j}{N}.
\end{equation}
With this privacy definition, we have adjusted the original fidelity evaluation \cite{lucic2018gans} shown in eq. (\ref{eq:fidelity}) to a privacy violation considered formula $f^p$ in equation \ref{eq:pfidelity}, 
\begin{equation}\label{eq:fidelity}
f_j=
\left\{ 
  \begin{array}{ c l }
    0 & \quad \textrm{if } \exists i\in [0,M) \to s_j\in S_i^{\rm copy} \cup S_i^{\rm real} \\
    1                 & \quad \textrm{otherwise}
  \end{array}
\right. \textrm{and} \quad  F=\frac{\sum_0^{N}f_j}{N};
\end{equation}
\begin{equation}\label{eq:pfidelity}
f^p_j=
\left\{ 
  \begin{array}{ c l }
    0 & \quad \textrm{if } \exists i\in [0,M) \to s_j\in S_i^{\rm real} \\
    1                 & \quad \textrm{otherwise}
  \end{array}
\right.\textrm{and} \quad F^p=\frac{\sum_0^{N}f^p_j}{N}.
\end{equation}
The measurements of image distance can be tricky due to the high resolution of the original images (512×512). Thus, we first used VQ-VAE \cite{van2017neural,razavi2019generating} to quantize all images to integral latent feature maps, i.e., each pixel in the latent feature maps is a $Q$-way categorical variable, sampling from 0 to $Q-1$, and then compute the Hamming distance between these images. In our experiments, we used $Q=256$. \\

\noindent \textbf{Variety. }To measure the variety, we introduced the JPEG file size of the mean image. The lossless JPEG file size of the group average image was used to measure the inner class variety in the ImageNet dataset \cite{deng2009imagenet}. This approach was justified by the authors who presumed that a dataset containing diverse images would result in a blurrier average image, thus reducing the lossless JPEG file size of the mean image. To ensure that the variety score is positively correlated with the true variety, we normalized it to $[0,1]$ across all groups of synthetic images, and then subtracted it from 1 to obtain the final variety score. It is worth noting that variety can also be quantified by the standard deviation of the discrete latent features. However, in our study, we chose to measure variety in the original image space to better align with human perception.\\

\noindent\textbf{Utility. }We divided our X-ray dataset into four groups: training datasets A1 and A2, validation set B, and testing set C. We also included an additional open-access pediatric X-ray dataset, D. For our simulation, we treated A1 as a local dataset and A2 as a remote dataset that cannot be accessed by A1. We evaluated the utility of synthetic data in two conditions:
\begin{enumerate}
    \item A1 vs. adding synthetic data generated from A1. In this condition, no privacy issue is considered. 
    \item A1 vs. adding synthetic data generated from A2. In this condition, synthetic data will be evaluated using privacy protection skills. 
\end{enumerate}
\indent Under both conditions, we evaluated the intra-task augmentation utility and cross-task augmentation utility to simulate real-world use cases for synthetic data. Intra-task augmentation utility is measured by the percentage improvement in classification accuracy of C when adding synthetic data to the training dataset. We used a paired Wilcoxon signed-rank test to assess the significance of the accuracy improvement. If the improvement is significant, it indicates that the synthetic images are useful. We compared the augmentation utility with simple augmentations, such as random flipping, rotating, and contrasting. \\
\indent The cross-task augmentation utility is determined by the power of features extracted from the models trained with synthetic data. We used the models to extract features from D and trained a Support Vector Machine classifier on these features to measure accuracy. This allowed us to evaluate whether synthetic images can provide powerful features that facilitate downstream tasks. Similarly, the cross-task augmentation utility is also the percentage improvement in classification accuracy compared to the model trained only on A1. 

\section{Experimental Settings and Parameters}
We primarily evaluated the performance of synthetic data on the CheXpert dataset, with a focus on identifying the presence of pleural effusion (PE). To perform our evaluation, we split the large dataset into four subsets: A1 (15004 with PE and 5127 without PE), A2 (30692 with PE and 10118 without PE), B (3738 with PE and 1160 PE), and C (12456 with PE and 3863 without PE). To evaluate the cross-task utility of synthetic models, we used an X-ray dataset D \footnote{\url{https://www.kaggle.com/datasets/paultimothymooney/chest-xray-pneumonia?resource=download}} consisting of 5863 images of pediatric patients with pneumonia and normal controls. We resized all X-ray images to a resolution of 512×512 before evaluation.\\
\indent For the StyleGAN2 method, we utilized six truncation parameters during sampling to generate six sets of synthetic images ($\phi \in [0.0,0.2,0.4,0.6,0.8,1.0]$). In total, we trained 16 classification models on different combinations of datasets, including A1, A1+A2, A1+StyleGAN2-synthesized A1 (6 models), A1+LDM-synthesized A1, A1+StyleGAN2-synthesized A2 (6 models), and A1+LDM- synthesized A2. For further information on implementation details, hyperparameters and table values, please refer to our supplementary file and our publicized codes \url{https://github.com/ayanglab/MedSynAnalyzer}.

\section{Experimental Results}
% The detailed values will be presented in our supplementary file. In this section, we will only present diagrams derived from the original data to better demonstrate our conclusions. 
\subsection{The Proposed Metrics Match with Human Perception}	
In our work, we proposed to use VQ-VAE to extract discrete features from original high-resolution X-ray images. To prove the validity of our method, we selected twenty images from each synthetic dataset and dataset A1 and invited two clinicians to rate the fidelity and variety manually. The human perceptual fidelity is rated from 0 to 1; and the human perceptual variety is computed by the percentage of different scans identified from the selected twenty synthetic images, i.e., if they thought all twenty patients were derived from the same scan, the human perceptual variety score is $1/20=0.05$. 
To assure a fair comparison, we allow discussion between them. The result is shown in Fig. \ref{fig:function} (c). The fidelity and variety score calculated with our method matched perfectly with human perception ($p<0.05$), and FID, which was highly influenced by mode collapse and increased over the diversity, failed to provide a valid analysis of image fidelity. 
\subsection{The Trade-off between Fidelity and Variety }	
All of our experiments showed a strong negative correlation between variety and fidelity, with a Pearson’s correlation coefficient of -0.92 ($p<0.01$). As it is widely known in GAN-based models, fidelity and variety are in conflict \cite{kynkaanniemi2019improved}. In this study, we further validated this by introducing the LDM model and demonstrating empirically that deep generative models inevitably face the trade-off between variety and fidelity (as shown by the grey lines in Fig 3 (a-b)).
\begin{figure}
    \centering
    \vspace{-6mm}
    \includegraphics[width=\textwidth]{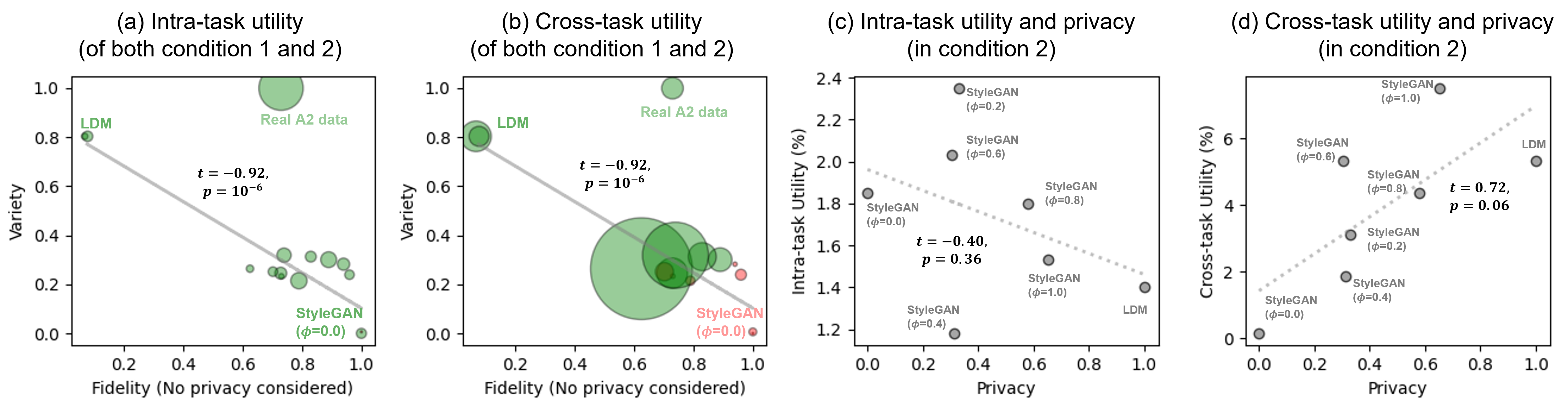}
    \vspace{-6mm}
    \caption{The relationships between fidelity, variety, and utility of adding additional synthetic data. The sizes of points in (a) and (b) are the improvement brought by adding synthetic data compared to no synthetic data being added. It should be noted that we visualized the point size in a power of 7 to better compare the improvements. Green points indicate significance using Signed Ranked T-test, and red indicates no significance. }
    \label{fig:correlation}
    \vspace{-6mm}
\end{figure}
\subsection{What Kind of Synthetic Data is Desired by Downstream Tasks When Privacy is Not an Issue?}	
\begin{figure}
    \centering
    \vspace{-10mm}
    \includegraphics[width=\textwidth]{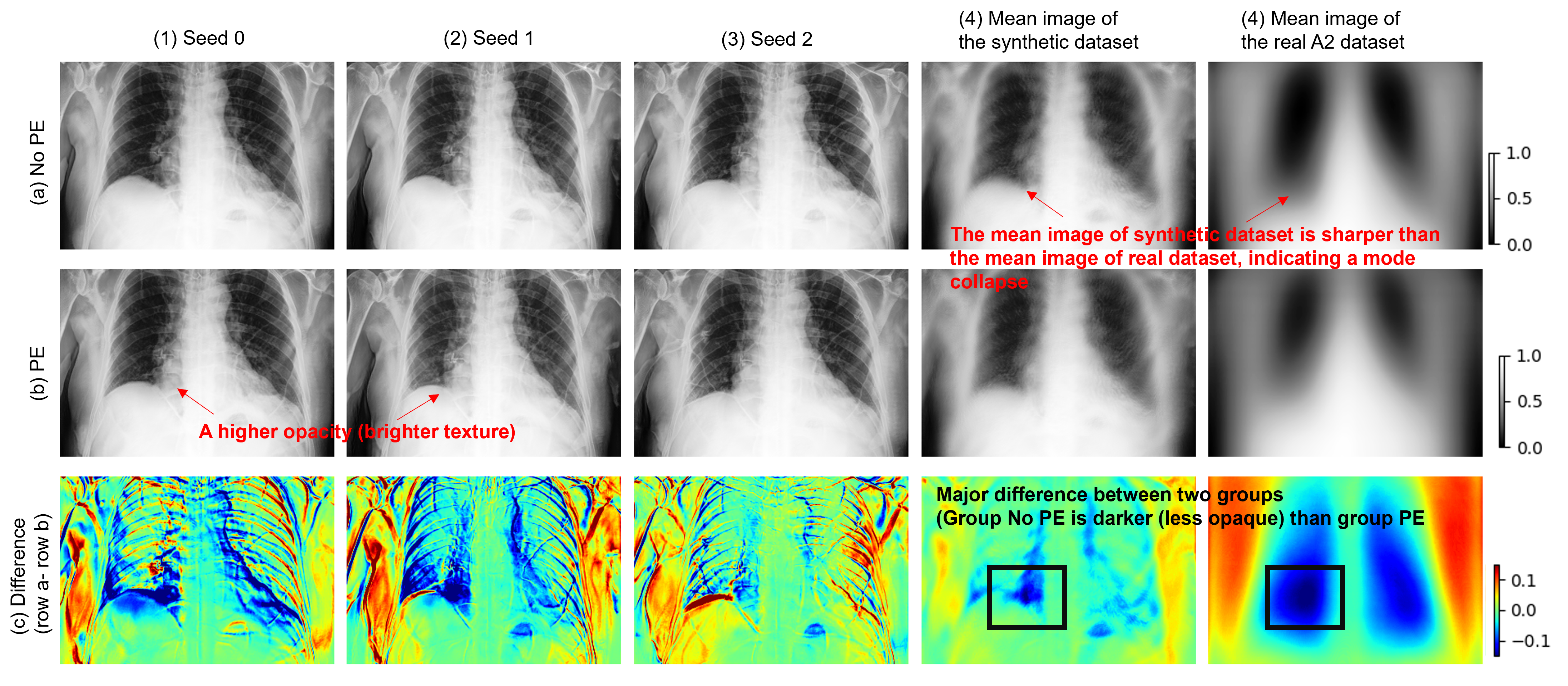}
    \vspace{-6mm}
    \caption{Example synthetic images sampled from the StyleGAN2 model with a truncation $\phi=0.2$ trained on A2 with different sampling seeds (1-3), as well as the mean image of the two groups calculated on the entire synthetic dataset (4). All the images in (a-b) were normalized to [0,1], and the difference between row (a) and row (b) is shown in row (c). The presence of PE in X-ray images is characterized by increased opacity of the lung near the hemidiaphragm, which matches the group difference of mode-collapsed images. Therefore, even in the case of severe mode collapse during synthesis, these synthetic images can still contribute to the improvement of intra-task classification accuracy.}
    \label{fig:diff}
    \vspace{-6mm}
\end{figure}

In this study, we cannot find a significant correlation  of utility between neither dimension (fidelity, variety nor privacy), indicating that there is currently no way to measure the utility except for large-scale experiments. However, we did observe a similar pattern of utility in our experiments shown in Fig. \ref{fig:correlation} (a-b). \\
\indent First, the intra-task augmentation utility favors synthetic data with higher fidelity (i.e., high $F$), even under mode collapses. For instance, when $\phi=0.2$ for StyleGAN, a mode collapse was observed. The model is producing images that look similar to each other, resulting in a mean image with a sharp contrast (Fig. \ref{fig:diff} (a4)). However, this mode collapse seems to highlight the difference between the presence and non-presence of PE (Fig. \ref{fig:diff} (c)), leading to a performance improvement in intra-task augmentation. The PE in X-ray images is fluid in the lowest part of the chest, forming a concave line obscuring the costophrenic angle and part or all of the hemidiaphragm. These opacity differences were highlighted in the mode-collapsed images, which, on the other hand, improved the classification accuracy of PE identification.\\  

\indent For cross-task augmentation, synthetic data with a higher variety is favored for its utility as mode collapse can limit the focus of classification networks and lead to poor generalization performance on other tasks. For instance, a network trained to focus on lung opacity differences near the hemidiaphragm may not help in the accurate diagnosis of pediatric pneumonia, which is the motivation behind dataset D. As shown in Fig. \ref{fig:correlation} (b), synthetic data with low variety but high fidelity is unable to contribute to powerful feature extraction. Therefore, variety in synthetic data is crucial for effective cross-task augmentation. \\
\indent As mentioned, we invited two radiologists to visually assess synthetic images. The visual inspection showed that all twenty LDM-synthesized images were easily recognized as fake due to their inability to capture the texture of X-ray images, as shown in the supplementary file. However, the shape and boundaries of the lungs were accurately captured by LDM. Despite their low visual fidelity, we demonstrated that these synthetic images still contribute to powerful feature extraction, which is crucial for cross-task utility.

\subsection{What Kind of Synthetic Data is Desired by Downstream Tasks When Privacy is an Issue?}
It is also discussed in the literature about the dilemma between utility and privacy \cite{stadler2022synthetic}. We reached a similar conclusion, i.e., there is a trade-off between utility and privacy in intra-class classification tasks, for which fidelity is considered to be crucial. Thus, a shift in the image domain could lead to a decrease in intra-task utility. 

However, as shown in Fig. \ref{fig:diff} (d), we observed that privacy and utility are not always conflicting. As discussed earlier, for cross-task augmentation, the utility favors synthetic images with high variety rather than high fidelity. Therefore, we demonstrated that it is possible to achieve both privacy and utility in cross-task augmentation scenarios. Fig. \ref{fig:diff} (d) shows an interesting positive correlation between privacy and cross-task utility. However, it is important to note that this does not imply a causal relationship between privacy and cross-task utility. Rather, the positive correlation is caused by the mode collapse during synthesis. Mode collapse can lead to a lack of diversity in generated data, which in turn can make it easier to identify individuals or sensitive information in the generated data, i.e., mode collapses are more likely to result in a high possibility of privacy breach as well as low cross-task utility.

\section{Conclusion}
In this work, we proposed a four-dimensional evaluation metric for synthetic images, including a novel privacy evaluation score and utility evaluation score. Through intensive experiments in over 100k chest X-ray images, we drew three major conclusions which we can envision that have broad applicability in medical image synthesis and analysis.

Firstly, there is an inevitable trade-off among different aspects of synthetic images, especially between fidelity and variety. Secondly, different downstream tasks require different properties of synthetic images, and synthetic images do not necessarily have to reach high metric scores across all aspects to be useful. Traditionally, low fidelity and mode collapses have been treated as disadvantages in data synthesis, and numerous algorithms have been proposed to fix these issues. However, our work demonstrates that these failures of synthetic data do not always sabotage their utility as expected.
Lastly, we have showed that it is possible to achieve both privacy and utility in transfer learning problems.

In conclusion, our work contributes to the development of synthetic data as a valuable solution to enrich real-world datasets, to evaluate thoroughly medical image synthesis as a pathway to overall enhance medical image analysis tasks.

% \subsubsection{Acknowledgements} Please place your acknowledgments at
% the end of the paper, preceded by an unnumbered run-in heading (i.e.
% 3rd-level heading).

\bibliographystyle{splncs04}
\bibliography{ref}

\end{document}